\documentclass[fleqn]{new_tlp}

\usepackage[utf8]{inputenc}
\usepackage{mathptmx}
\usepackage[lighttt]{lmodern} 
\usepackage{microtype}
\usepackage{amsmath,amssymb}
\usepackage{url}
\usepackage{caption}
\usepackage{graphicx}
\usepackage[x11names]{xcolor}
\usepackage{tikz}
\usetikzlibrary{arrows,shapes}
\usepackage{listings}
\newtheorem{example}{Example}
\newcommand{\eofex}{\hfill$\blacksquare$}

\newcommand{\hreduce}{\hspace*{-1.7mm}}
\newcommand{\kreduce}{\hspace{-0.65pt}}
\newcommand{\ereduce}{} 
\newcommand{\freduce}{} 
\newcommand{\preduce}{} 

\newcommand{\clingo}{\emph{clingo}}
\newcommand{\openTCS}{\emph{openTCS}}

\allowdisplaybreaks

\title[Routing Driverless Transport Vehicles in Car Assembly with ASP]%
{Routing Driverless Transport Vehicles in Car Assembly with Answer Set Programming}%

\author[M.~Gebser,
  P.~Obermeier,
  M.~Ratsch-Heitmann,
  M.~Runge,
  T.~Schaub]%
{%
  Martin Gebser,
  Philipp Obermeier,
  Torsten Schaub\\
  University of Potsdam, Germany
  \and
  Michel Ratsch-Heitmann,
  Mario Runge\\
  Mercedes-Benz Ludwigsfelde GmbH, Germany}

\submitted{[n/a]}
\revised{[n/a]}
\accepted{[n/a]}

\begin{document}

\maketitle

\begin{abstract}
Automated storage and retrieval systems are principal components of modern
production and warehouse facilities.
In particular, automated guided vehicles nowadays substitute human-operated
pallet trucks in transporting production materials between storage locations
and assembly stations.
While low-level control systems take care of navigating such driverless vehicles along
programmed routes and avoid collisions even under unforeseen circumstances,
in the common case of multiple vehicles sharing the same operation area,
the problem remains how to set up routes such that a collection of transport tasks
is accomplished most effectively.
We address this prevalent problem in the context of car assembly at
Mercedes-Benz Ludwigsfelde GmbH, a large-scale producer of commercial vehicles,
where routes for automated guided vehicles used in the production process
have traditionally been hand-coded by human engineers.
Such ad-hoc methods may suffice as long as a running production process
remains in place, while any change in the factory layout or production targets
necessitates tedious manual reconfiguration,
not to mention the missing portability between different production plants.
Unlike this, we propose a declarative approach based on Answer Set Programming
to optimize the routes taken by automated guided vehicles for accomplishing
transport tasks.
The advantages include a transparent and executable problem formalization,
provable optimality of routes relative to objective criteria, as well as
elaboration tolerance towards particular factory layouts and production targets.
Moreover, we demonstrate that our approach is efficient enough to deal with the
transport tasks evolving in realistic production processes at the car factory of
Mercedes-Benz Ludwigsfelde GmbH.

\medskip\noindent
{\em Under consideration for publication in Theory and Practice of Logic Programming (TPLP)}
\end{abstract}



\section{Introduction}\label{sec:introduction}

Automated guided vehicles play a key role in modern industries,
let it be in warehouses, mines, or as in our case production facilities.
Most of the time, however,
these vehicles are programmed by human engineers to execute specific tasks.
This makes it impossible to quickly reassign tasks in case of breakdowns
or to easily react to changing production requirements,
not to mention the missing portability between different production plants and factory layouts.
The lack of elaboration tolerance does not only lead to high expenditures,
but the resulting rigid control also 
rules out any flexible fleet management.
In particular, no conclusions can be drawn about the effectiveness or even optimality
of pre-programmed vehicle routes. 

\begin{figure}[t]
\centering
\includegraphics
[scale=0.224]{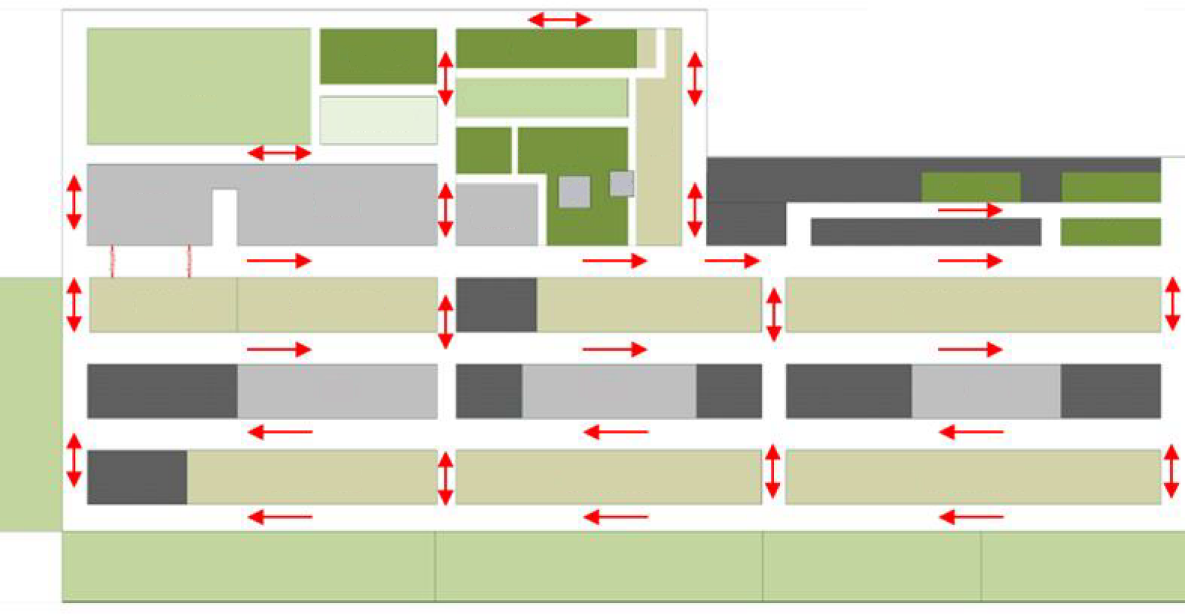}
\caption{Real-world factory layout with transport corridors and directions indicated by arrows\label{fig:factory}}
\end{figure}

In view of these common circumstances,
we address the challenge of devising a new control system for automated guided vehicles
supplying the assembly lines at the car factory of Mercedes-Benz Ludwigsfelde GmbH,
which is expected to be
flexible enough to adapt to malfunctions and emerging requirements, and
whose quality of operation can be measured and optimized
relative to given objectives.
To make our specific task more precise,
consider the retouched layout of an assembly hall at the production plant of Mercedes-Benz Ludwigsfelde GmbH
in Figure~\ref{fig:factory}.
%
The overall goal is to guarantee that all car components are at their designated place next to the assembly line
when they are due for installation.
The corresponding transport tasks are accomplished by a heterogeneous fleet of automated guided vehicles that fetch the necessary components from
several storage areas.
For instance, a vehicle may first halt at some storage location to load production material,
then move on to an assembly station in need of the material, and from there
cart off leftover material to a recycling facility.
The sketched task thus involves stopovers at three distinct locations,
between which the vehicle must pick a route without getting blocked by others,
where dedicated parking spaces are included in the layout to let vehicles make room.
Notably, the regular production process runs periodically, so that the
given transport tasks have to be repeatedly executed in fixed intervals.

To make the execution of transport tasks more effective,
we propose to perform task assignment and vehicle routing
by means of Answer Set Programming (ASP;~\citeNP{lifschitz99b}),
given its declarative and elaboration tolerant approach to combinatorial multi-objective problem solving.
This results in a transparent and easily adjustable problem description
along with 
optimal solutions relative to objective criteria.
Regarding performance,
it also turns out that our approach is efficient enough to
meet the industrial-scale 
requirements of production processes at the car factory of Mercedes-Benz Ludwigsfelde GmbH.

The paper is structured as follows.
In the next section, we start by formalizing 
automated guided vehicle routing for transport tasks 
evolving in production processes at the car factory of Mercedes-Benz Ludwigsfelde GmbH.
Once this is accomplished,
we provide in Section~\ref{sec:encoding} a corresponding ASP encoding,
incorporating the hard and soft constraints on solutions.
In Section~\ref{sec:experiments},
we empirically evaluate our approach on use cases designed to test the
practical applicability of control systems for automated guided vehicles.
Finally, we conclude the paper with a brief discussion of related work
and the achieved results.




\section{Problem Formalization}\label{sec:approach}

In what follows, we formally describe the automated guided vehicle routing scenarios
faced in production processes at the car factory of Mercedes-Benz Ludwigsfelde GmbH,
the conditions on their solutions, as well as
objective criteria concerning solution quality devised
together with production engineers at Mercedes-Benz Ludwigsfelde GmbH.

An automated guided vehicle routing scenario
is specified in terms of a directed graph $(V,E)$,
where the nodes~$V$ stand for \emph{locations} of interest and the edges $E\subseteq V\times V$
provide \emph{connections} between them, along with a set~$T$ of transport \emph{tasks}.
Among the nodes in~$V$, we distinguish particular \emph{halt} and \emph{park} nodes,
given by $h(V)\subseteq V$ and $p(V)\subseteq V$
such that $h(V)\cap p(V)=\emptyset$ holds.
Moreover, each task $t\in T$ has an associated non-empty sequence $\langle s_1,\dots,s_m \rangle$ of
\emph{subtasks}, whose elements are halt nodes, i.e., $\{s_1,\dots,s_m\}\subseteq h(V)$,
and we write $s(t)=m$ and $t[i]=s_i$, for $1\leq i\leq m$, to refer to the
number of or individual subtasks, respectively, associated with~$t$.
In addition, every edge, halt or park node, and task is characterized by some positive integer,
denoted by $d(x)$ for $x\in E\cup h(V)\cup p(V)\cup T$, where
$d(e)$ expresses the \emph{move duration} for a connection $e\in E$,
$d(v)$ provides the \emph{halt} or \emph{park duration} for a location $v\in h(V)\cup p(V)$, and
$d(t)$ constitutes the \emph{deadline} for completing a task $t\in T$.
Finally, a set~$C$ of \emph{vehicles} usable for transport tasks
is given together with, for each $c\in C$, an initial location $l(c)\in V$
such that $l(c)\neq l(c')$ when $c'\in C\setminus\{c\}$.
The vehicles can accomplish transport tasks by taking connections
and halting at the locations of respective subtasks,
whose halt durations reflect time spent to operate carried materials
(e.g., loading or unloading),
while further stops are admitted at park nodes only,
included for allowing a vehicle to wait in case its subsequent route is temporarily blocked by others.

\begin{figure}[t]
\centering
\begin{tikzpicture}[->,>=stealth',auto,node distance=1.5cm,thick]
\tikzset{base/.style={draw,inner sep=0pt,minimum size=0.7cm,circle}}
\tikzset{halt/.style={draw,inner sep=0pt,minimum size=0.8cm,diamond}}
\tikzset{park/.style={draw,inner sep=0pt,minimum size=0.6cm,rectangle}}
\tikzset{pure/.style={}}
\node[halt] (2) {$2$};
\node[base] (3) [above of=2,yshift=1.5cm] {$3$};
\node[base] (1) [right of=2,xshift=1.5cm] {$1$};
\node[halt] (4) [right of=3,xshift=1.5cm] {$4$};
\node[halt] (6) [right of=1,xshift=1.5cm] {$6$};
\node[halt] (5) [right of=4,xshift=1.5cm] {$5$};
\node[park] (7) [above of=1] {$7$};
\node[pure] (v1) [below of=1,yshift=0.9cm] {\textcolor{Red2}{$c_1:0$}};
\node[pure] (v2) [below of=2,yshift=0.9cm] {\textcolor{Blue1}{$c_2:0$}};
\node[pure] (t13) [left of=2,xshift=0.4cm,yshift=0.2cm] {\textcolor{Red2}{\small{$t_1[3]:55$}}};
\node[pure] (t23) [left of=2,xshift=0.4cm,yshift=-0.2cm] {\textcolor{Blue1}{\small{$t_2[3]:49$}}};
\node[pure] (t12) [above of=4,xshift=-0.7cm,yshift=-0.85cm] {\textcolor{Red2}{\small{$t_1[2]:40$}}};
\node[pure] (t22) [above of=4,xshift=0.7cm,yshift=-0.85cm] {\textcolor{Blue1}{\small{$t_2[2]:34$}}};
\node[pure] (t11) [right of=5,xshift=-0.4cm] {\textcolor{Red2}{\small{$t_1[1]:17$}}};
\node[pure] (t21) [right of=6,xshift=-0.4cm] {\textcolor{Blue1}{\small{$t_2[1]:19$}}};
\node[pure] (t1p) [right of=7,xshift=-1cm] {\textcolor{Red2}{\small{$6$}}};
\path (1) edge (2);
\path (1) edge (7);
\path (2) edge (3);
\path (3) edge (4);
\path (4) edge (5);
\path (4) edge (7);
\path (5) edge (6);
\path (6) edge (1);
\path (7) edge (1);
\path (7) edge (4);
\path (1) edge [transform canvas={xshift=4mm},color=Red2,shorten <=0.5mm] node [xshift=0.7mm] {\small{$4$}} (7);
\path (7) edge [transform canvas={xshift=4mm},color=Red2] node [xshift=0.9mm] {\small{$10$}} (4);
\path (4) edge [transform canvas={yshift=-4mm},color=Red2] node [yshift=-0.5mm] {\small{$14$}} (5);
\path (5) edge [transform canvas={xshift=-4mm},color=Red2] node [xshift=-0.8mm] {\small{$21$}} (6);
\path (6) edge [transform canvas={yshift=4mm},color=Red2,shorten >=0.5mm] node [yshift=0.5mm] {\small{$25$}} (1);
\path (1) edge [transform canvas={yshift=4mm},color=Red2,shorten <=0.5mm] node [yshift=0.5mm] (v112) {\small{$29$}} (2);
\path (2) edge [transform canvas={xshift=4mm},color=Red2,shorten >=0.5mm] node [xshift=0.9mm] {\small{$33$}} (3);
\path (3) edge [transform canvas={yshift=-4mm},color=Red2,shorten <=0.5mm] node [yshift=-0.5mm] {\small{$37$}} (4);
\path (4) edge [transform canvas={xshift=-4mm},color=Red2] node [xshift=-0.9mm] {\small{$44$}} (7);
\path (7) edge [transform canvas={xshift=-4mm},color=Red2,shorten >=0.5mm] node [xshift=-0.9mm] {\small{$48$}} (1);
\node[pure] [above of=v112,yshift=-7.5mm] {\textcolor{Red2}{\small{$52$}}};
\path (2) edge [transform canvas={xshift=-4mm},color=Blue1] node [xshift=0.5mm] {\small{$4$}} (3);
\path (3) edge [transform canvas={yshift=4mm},color=Blue1] node [yshift=-0.5mm] {\small{$8$}} (4);
\path (4) edge [transform canvas={yshift=4mm},color=Blue1] node [yshift=-0.5mm] {\small{$12$}} (5);
\path (5) edge [transform canvas={xshift=4mm},color=Blue1] node [xshift=-0.5mm] {\small{$16$}} (6);
\path (6) edge [transform canvas={yshift=-4mm},color=Blue1] node [yshift=0.5mm] {\small{$23$}} (1);
\path (1) edge [transform canvas={xshift=6.5mm},color=Blue1,shorten <=0.75mm] node [xshift=5.5mm] {\small{$27$}} (7);
\path (7) edge [transform canvas={xshift=6.5mm},color=Blue1,shorten >=0.25mm] node [xshift=5.5mm] {\small{$31$}} (4);
\path (4) edge [transform canvas={xshift=-6.5mm},color=Blue1,shorten <=0.25mm] node [xshift=-5.5mm] {\small{$38$}} (7);
\path (7) edge [transform canvas={xshift=-6.5mm},color=Blue1,shorten >=0.75mm] node [xshift=-5.5mm] {\small{$42$}} (1);
\path (1) edge [transform canvas={yshift=-4mm},color=Blue1] node [yshift=0.5mm] {\small{$46$}} (2);
\end{tikzpicture}

\caption{Optimal routes to accomplish two transport tasks with three subtasks each by two vehicles\label{fig:instance}}
\end{figure}
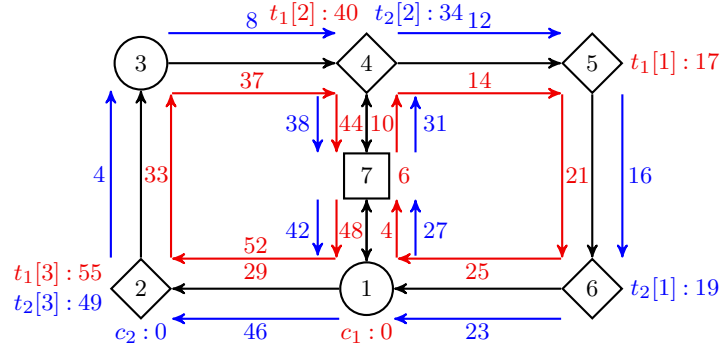

\begin{example}\label{ex:instance}
The scenario depicted (in black) in Figure~\ref{fig:instance}
consists of the nodes $V=\{1,\dots,7\}$ and edges
$E=\{(1,2),\linebreak[1](1,7),\linebreak[1](2,3),\linebreak[1](3,4),\linebreak[1]
     (4,5),\linebreak[1](4,7),\linebreak[1](5,6),\linebreak[1](6,1),\linebreak[1](7,1),\linebreak[1](7,4)\}$.
The halt nodes $h(V)=\{2,4,5,6\}$ are indicated by diamonds in Figure~\ref{fig:instance},
and the park node in $p(V)=\{7\}$ is marked by a square.
The two tasks in $T=\{t_1,t_2\}$ include three subtasks each, where
the respective sequences of halt nodes,
$\langle t_1[1]=5,\linebreak[1]t_1[2]=4,\linebreak[1]t_1[3]=2 \rangle$ and
$\langle t_2[1]=6,\linebreak[1]t_2[2]=4,\linebreak[1]t_2[3]=2 \rangle$,
are listed within red or blue labels, respectively, near the subtasks.
Similarly, the initial locations $l(c_1)=1$ and $l(c_2)=2$ of the
vehicles in $C=\{c_1,c_2\}$ are indicated by corresponding labels.
While Figure~\ref{fig:instance} does not display the durations,
we assume a uniform move    duration $d(e)=4$ per connection $e\in E$,
a halt duration $d(v)=3$ at each of the four halt nodes $v\in \{2,4,5,6\}$,
and a park duration $d(7)=2$ at the park node~$7$.
Moreover, the deadline for both tasks, $t_1$ and $t_2$, is given by
$d(t_1)=d(t_2)=60$ and also omitted in Figure~\ref{fig:instance} for better clarity.
\eofex
\end{example}

A solution to a guided vehicle routing scenario comprises a \emph{task assignment}
$\alpha:T \rightarrow C$ along with a strict partial \emph{order} ${\prec}$ on $T$
such that either $t\prec t'$ or $t'\prec t$ holds when $\alpha(t)=\alpha(t')$ for
tasks $t\neq t'$.
In addition, for each vehicle $c\in C$, it contains a \emph{route}
$\pi(c) = \langle r_1,\dots,r_k \rangle$,
where $\{r_1,\dots,r_k\}\subseteq E\cup h(V)\cup p(V)$, subject to the following
conditions:
\begin{enumerate}
\item If $r_i=(v,v')$, for $(v,v')\in E$, or $r_i=v$, for $v\in h(V)\cup p(V)$,
      holds for some $1\leq i\leq k$,
      then we require that $r_{i-1}=(v'',v)$ or $r_{i-1}=v$,
      where we let $r_0=l(c)$.
\item For any subtask $t[j]$ of a task $t\in T$ with $\alpha(t)=c$,
      we define its completion index by
      $g(t[j])=\min\{i \mid 1\leq i \leq k, r_i=t[j],
                            \max(\{g(t[j-1])    \mid 1 < j\}\cup
                                 \{g(t'[s(t')]) \mid t' \prec t\}) < i\}$
      and require that $g(t[j])\leq \sum_{1\leq i\leq g(t[j])}d(r_i)\leq d(t)$.
      In words, the subtasks of all tasks assigned to a vehicle~$c$ have to be completed
      in the order given by $\prec$ (as well as sequences of subtasks) within their respective deadlines,
      where a subtask is completed once the route $\pi(c)$ includes
      a halt at its location such that all preceding subtasks have been completed before.
\item If $r_i=v$ for a halt node $v\in h(V)$ and $1\leq i\leq k$,
      then we require that $g(t[j])=i$ for some subtask $t[j]$
      of a task $t\in T$ with $\alpha(t)=c$.
      That is, halts may be included in the route of a vehicle~$c$
      exclusively for completing subtasks of tasks assigned to~$c$.
\item With each $x\in E\cup V$,
      we associate a set of occupation times defined as
      $u(c,x)=\{d+\sum_{1\leq j< i}d(r_j) \mid 1\leq i \leq k,r_i=x,1\leq d\leq d(x)\}
              \cup
              \{\sum_{1\leq j\leq i}d(r_j) \mid 1\leq i \leq k,r_i\in E,\linebreak[1]r_i=(v,x)\}$.
      For any vehicle $c' \in C\setminus\{c\}$, the following requirements check that
      the routes of~$c$ and~$c'$ do not both lead to a joint location at the same time,
      and that~$c$ and~$c'$ do not meet in between nodes by taking connections in opposite directions:
      for each location $v\in V$, we require that $u(c,v)\cap u(c',v) = \emptyset$,
      while $u(c,(v,v'))\cap u(c',(v',v)) = \emptyset$ must hold for each
      bidirectional connection pair $\{(v,v'),(v',v)\}\subseteq E$.
\end{enumerate}

\begin{example}\label{ex:solution}
The red and blue arrows and labels in Figure~\ref{fig:instance}
indicate the routes of
a solution to the automated guided vehicle routing scenario from Example~\ref{ex:instance},
where the task assignment is given by $\alpha(t_1)=c_1$ and $\alpha(t_2)=c_2$.
As no distinct tasks are assigned to the same vehicle,
we have that the order relation ${\prec}$ is empty,
while $c_1$ and $c_2$ have to complete the subtasks of $t_1$ or~$t_2$, respectively,
in order.
The routes taken by $c_1$ and $c_2$ can be traced by
considering the times of completing route elements,
included in labels along nodes and edges, in increasing order,
starting from $c_1:0$ and $c_2:0$ at the initial locations
$l(c_1)=1$ and $l(c_2)=2$.
For vehicle~$c_1$, this yields the route
$\pi(c_1)=\langle (1,7),\linebreak[1]7,\linebreak[1](7,4),\linebreak[1](4,5),\linebreak[1]5,\linebreak[1]
                  (5,6),\linebreak[1](6,1),\linebreak[1](1,2),\linebreak[1](2,3),\linebreak[1](3,4),\linebreak[1]
                  4,\linebreak[1](4,7),\linebreak[1](7,1),\linebreak[1](1,2),\linebreak[1]2 \rangle$,
where the four stops at halt or park nodes in $h(V)\cup p(V) = \{2,4,5,6,7\}$ are of particular interest.
In fact, the inclusion of the park node~$7$ at the second position of $\pi(c_1)$ delays the
subsequent move $(7,4)$ by the park duration $d(7)=2$ in order to avoid arriving at node~$4$
at the same time as the other vehicle~$c_2$.
Unlike that, the later stops at the halt nodes $5$, $4$, and~$2$ are made to complete
the sequence $\langle t_1[1]=5,\linebreak[1]t_1[2]=4,\linebreak[1]t_1[3]=2 \rangle$
of subtasks.
While $\pi(c_1)$ makes $c_1$ revisit the halt nodes $t_1[2]=4$ and $t_1[3]=2$,
a halt would not be admitted on the first arrival because the
preceding subtask $t_1[1]=5$ or $t_1[2]=4$, respectively, has not yet been completed. 
Hence, the halt at~$5$ comes before returning to and halting at~$4$,
and then in turn at~$2$ at the very end of the route~$\pi(c_1)$.
As one can check, the sum of durations over connections as well as halt and park nodes
in $\pi(c_1)$ is $55$, which means that the task~$t_1$ gets completed within its deadline $d(t_1)=60$.
Regarding the other vehicle~$c_2$, tracing the blue labels in Figure~\ref{fig:instance}
yields the route
$\pi(c_2)=\langle (2,3),\linebreak[1](3,4),\linebreak[1](4,5),\linebreak[1](5,6),\linebreak[1]
                  6,\linebreak[1](6,1),\linebreak[1](1,7),\linebreak[1](7,4),4,\linebreak[1]\linebreak[1]
                  (4,7),\linebreak[1](7,1),\linebreak[1](1,2),\linebreak[1]2 \rangle$
for completing
the sequence $\langle t_2[1]=6,\linebreak[1]t_2[2]=4,\linebreak[1]t_2[3]=2 \rangle$
of subtasks.
Along this route, $c_2$ visits the halt nodes $t_2[2]=4$ and $t_2[3]=l(c_2)=2$ twice,
where halts are in both cases made on the second arrival for completing the subtasks of~$t_2$ in order.
Moreover, no stop at the park node~$7$ is needed to avoid meeting the other vehicle~$c_1$,
and the durations of connections and halt nodes in $\pi(c_2)$ sum up to~$49$,
so that also the task~$t_2$ gets completed within its deadline $d(t_2)=60$.
\eofex
\end{example}

While an automated guided vehicle routing scenario may have plenty feasible solutions,
we apply the following objective criteria, below ordered by significance,
to distinguish preferred collections of routes $\pi(c_i)=\langle r_{1_i},\dots,r_{k_i} \rangle$
for the vehicles $c_i\in C$:

\begin{enumerate}
\item The \emph{makespan} $\mathit{ms}$ of a solution is the maximum sum of durations
      over the route elements of some vehicle, i.e.,
\begin{equation*} 
      \mathit{ms}=\max\{\mbox{$\sum$}_{1\leq j_i\leq {k_i}}d(r_{j_i}) \mid c_i\in C\}\text{.}
\end{equation*} 
\item The \emph{route length} $\mathit{rl}$ of a solution is the sum of durations
      over elements in the routes of all vehicles, i.e.,
\begin{equation*} 
      \mathit{rl}=\mbox{$\sum$}_{c_i\in C,1\leq j_i\leq {k_i}} d(r_{j_i})\text{.}
\end{equation*} 
\item The \emph{crossing number} $\mathit{cn}$ is the number of pairs $(\{c_i,c_{i'}\},v)$
      such that $c_i\in C$, $c_{i'}\in C\setminus\{c_i\}$, and we have that
      $r_{j_i}=(v',v)$ and $r_{j_{i'}}=(v'',v)$ for some connections
      $\{(v',v),(v'',v)\}\subseteq E$ with $v'\neq v''$,
      $1\leq j_i\leq k_i$, and $1\leq j_{i'}\leq k_{i'}$, i.e.,
\ereduce
\begin{equation*}
      \mathit{cn}=|\{(\{c_i,c_{i'}\},v) \mid {}
      \begin{array}[t]{@{}l@{}}
        c_i\in C, c_{i'}\in C\setminus\{c_i\},
        1\leq j_i\leq k_i, 1\leq j_{i'}\leq k_{i'}, {}
      \\
        \{r_{j_i},r_{j_{i'}}\}\subseteq E,
        r_{j_i}=(v',v), r_{j_{i'}}=(v'',v), v'\neq v''\}|\text{.}
\freduce
      \end{array}
\end{equation*}
      In words, the crossing number estimates how often the routes of
      distinct vehicles come together at a joint location (at different times).
      Although solutions are such that vehicles do not share any location
      at the same time, it is desirable to make routes as disjoint as possible
      to keep knock-on effects in case of a disruption short.
\item The \emph{overlap number} $\mathit{on}$ is the number of triples $(\{c_i,c_{i'}\},r_{j_i},r_{j_{i'}})$
      such that $c_i\in C$, $c_{i'}\in C\setminus\{c_i\}$, and we have that
      $r_{j_i}=(v,v')$ and $r_{j_{i'}}=(v,v')$ (or $r_{j_{i'}}=(v',v)$)
      for some connection $(v,v')\in E$ (or connections $\{(v,v'),(v',v)\}\subseteq E$),
      $1\leq j_i\leq k_i$, and $1\leq j_{i'}\leq k_{i'}$, i.e.,
\ereduce
\begin{equation*}
      \mathit{on}=|\{(\{c_i,c_{i'}\},r_{j_i},r_{j_{i'}}) \mid {}
      \begin{array}[t]{@{}l@{}}
        c_i\in C, c_{i'}\in C\setminus\{c_i\},
        1\leq j_i\leq k_i, 1\leq j_{i'}\leq k_{i'}, {}
      \\
        \{r_{j_i},r_{j_{i'}}\}\subseteq E,
        r_{j_i}=(v,v'), r_{j_{i'}}\in\{(v,v'),(v',v)\}\}|\text{.}
\freduce
      \end{array}
\end{equation*}
      The motivation for aiming at a small overlap number is the same as
      with the crossing number, while connections taken by distinct vehicles
      are considered instead of crossing locations.
      Note that, for a bidirectional connection pair $\{(v,v'),(v',v)\}\subseteq E$,
      two vehicles $c_i$ and $c_{i'}$ that move between $v$ and $v'$ in either direction
      can contribute one, two, or four triples
      $(\{c_i,c_{i'}\},r_{j_i},r_{j_{i'}})$, depending on whether neither, one, or
      both include $(v,v')$ as well as $(v',v)$ in their routes.
      Unlike that, only the triple $(\{c_i,c_{i'}\},(v,v'),(v,v'))$ is obtained
      when $(v,v')\in E$ but $(v',v)\notin E$.
\end{enumerate}

For each of the four objective criteria specified above, a smaller value is preferred
to greater ones.
The order of significance is chosen such that the completion of all transport tasks
in as little time as possible has the highest priority,
minimizing the overall utilization of vehicles comes second,
keeping routes disjoint is third, and the
avoidance of overlapping connections takes the lowest priority.

\begin{example}\label{ex:optimal}
The solution to the automated guided vehicle routing scenario from Example~\ref{ex:instance}
indicated in Figure~\ref{fig:instance} and described further in Example~\ref{ex:solution} happens
to be optimal relative to the applied objective criteria.
Given that the sums $55$ and $49$ are obtained for the durations of elements of the route
$\pi(c_1)$ or $\pi(c_2)$, respectively,
the makespan $\mathit{ms}$ matches the maximum $55$,
and the route length $\mathit{rl}$ amounts to $55+49=104$.
Regarding the crossing number~$\mathit{cn}$, $\pi(c_1)$ and $\pi(c_2)$
visit three locations, $1$, $4$, and~$7$, via different connections,
as witnessed by occurrences of $(6,1)$ and $(7,1)$, $(3,4)$ and $(7,4)$, as well as
$(1,7)$ and $(4,7)$ in $\pi(c_1)$ or $\pi(c_2)$, respectively.
Moreover, the overlap number $\mathit{on}=14$ is obtained in view
of the inclusion of the connections $(1,2)$, $(2,3)$, $(3,4)$, $(4,5)$, $(5,6)$, and $(6,1)$
in both $\pi(c_1)$ and $\pi(c_2)$ along with moves between the nodes~$1$ and~$4$
as well as $4$ and~$7$ in either direction, each of the latter contributing four triples
counted together by $\mathit{on}$.
While the gap between the makespan $\mathit{ms}=55$ and the deadline
$d(t_1)=d(t_2)=60$ for $t_1$ and $t_2$ may seem small,
let us mention that 561 solutions are feasible and that
only one of them is optimal.
Such discrepancy clearly indicates that human engineers will hardly be able
to perform an exhaustive optimization of routes in realistic scenarios of greater size
only by hand.
\eofex
\end{example}


\section{Problem Encoding}\label{sec:encoding}

\begin{table}[t]
\centering
\hreduce
\begin{tabular}{|l|@{\!\!}l|}
\cline{1-2}
Fact & Meaning
\\\cline{1-2}
\mbox{\lstinline{node(}}$v$\mbox{\lstinline{)}}
&
$v\in V$
\\
\mbox{\lstinline{halt(}}$v$\mbox{\lstinline{,}}$d(v)$\mbox{\lstinline{)}}
&
$v\in h(V)$ with halt duration $d(v)$
\\
\mbox{\lstinline{park(}}$v$\mbox{\lstinline{,}}$d(v)$\mbox{\lstinline{)}}
&
$v\in p(V)$ with park duration $d(v)$
\\
\mbox{\lstinline{stay(}}$v$\mbox{\lstinline{,}}$d(v)$\mbox{\lstinline{)}}
&
$v\in h(V)\cup p(V)$ with halt or park duration $d(v)>1$
\\
\mbox{\lstinline{edge(}}$v$\mbox{\lstinline{,}}$v'$\mbox{\lstinline{,}}$d(v,v')$\mbox{\lstinline{)}}
&
$(v,v')\in E$ with move duration $d((v,v'))$
\\
\mbox{\lstinline{less(}}$v'$\mbox{\lstinline{,}}$v''$\mbox{\lstinline{,}}$v$\mbox{\lstinline{)}}
&
$\{(v',v),(v'',v)\}\subseteq E$ for lexicographically consecutive locations $v'< v''$
\\
\mbox{\lstinline{time(}}$n$\mbox{\lstinline{)}}
&
$0\leq n\leq \max\{d(t) \mid t\in T\}$
\\
\mbox{\lstinline{task(}}$t$\mbox{\lstinline{)}}
&
$t\in T$
\\
\mbox{\lstinline{task(}}$t$\mbox{\lstinline{,}}$d(t)$\mbox{\lstinline{)}}
&
$t\in T$ with deadline $d(t)$
\\
\mbox{\lstinline{tasks(}}$t$\mbox{\lstinline{,}}$t'$\mbox{\lstinline{)}}
&
$t\in T$ and $t'\in T\setminus\{t\}$
\\
\mbox{\lstinline{subtask(}}$t$\mbox{\lstinline{,s(}}$i$\mbox{\lstinline{))}}
&
$t\in T$ with $1\leq i\leq s(t)$
\\
\mbox{\lstinline{subtask(}}$t$\mbox{\lstinline{,s(}}$i$\mbox{\lstinline{),}}$v$\mbox{\lstinline{)}}
&
$t\in T$, $1\leq i\leq s(t)$, and $t[i]=v$ with $v\in h(V)$
\\
\mbox{\lstinline{vehicle(}}$c$\mbox{\lstinline{)}}
&
$c\in C$
\\
\mbox{\lstinline{vehicle(}}$c$\mbox{\lstinline{,}}$v$\mbox{\lstinline{)}}
&
$c\in C$ with initial location $l(c)=v$
\\\cline{1-2}
\end{tabular}
\caption{Fact format for specifying automated guided vehicle routing scenarios in ASP\label{tab:instance}}
\end{table}
Following the common modeling methodology of ASP,
we represent (optimal) solutions to automated guided vehicle routing scenarios
by facts specifying an instance along with a uniform problem encoding.
To begin with, Table~\ref{tab:instance} surveys the format of facts
describing the locations, connections, tasks, and vehicles belonging to
an automated guided vehicle routing scenario.
The respective fact representation of the scenario from Example~\ref{ex:instance}
is given in Listing~\ref{lst:instance},
using the shorthands `\lstinline{..}' and `\lstinline{;}'
of \clingo\ \cite{PotasscoUserGuide}
to abbreviate facts for a range or collection of arguments, respectively.
Note that facts of the form
\lstinline{stay(}$v$\lstinline{,}$d(v)$\lstinline{)}
combine halt and park nodes $v\in h(V)\cup p(V)$ whose associated duration
$d(v)$ is greater than one, as such non-atomic durations are subject to
dedicated conditions in our problem encoding below.
For locations $v\in V$ that have several incoming connections,
facts of the form
\lstinline{less(}$v'$\lstinline{,}$v''$\lstinline{,}$v$\lstinline{)}
provide lexicographically consecutive locations $v'<v''$ such that
$\{(v',v),(v'',v)\}\subseteq E$,
based on the standard term order of the ASP-Core-2 language \cite{aspcore2}.
Our encoding below makes use of the lexicographical order of predecessor locations
for a compact formulation of conditions to detect crossings. 
Moreover, facts
\lstinline{time(}$n$\lstinline{)}
give time points of interest, ranging from zero to the latest deadline of any
transport task,
and \lstinline{tasks(}$t$\lstinline{,}$t'$\lstinline{)} holds for distinct tasks
$t\in T$ and $t'\in T\setminus\{t\}$, which have to be
ordered by ${\prec}$ when $\alpha(t)=\alpha(t')$.
Facts of the remaining predicates list further properties of
automated guided vehicle routing scenarios in a straightforward way,
and their respective meanings are summarized in Table~\ref{tab:instance}.
\begin{lstlisting}[label=lst:instance,numbers=none,caption={Instance of automated guided vehicle routing with two tasks and two vehicles},float]
node(v(1..7)).
halt(v(2),3). stay(v(2),3).
halt(v(4),3). stay(v(4),3).
halt(v(5),3). stay(v(5),3).
halt(v(6),3). stay(v(6),3).
park(v(7),2). stay(v(7),2).

edge(v(6;7),v(1),4). less(v(6),v(7),v(1)).
edge(v(1),v(2),4).
edge(v(2),v(3),4).
edge(v(3;7),v(4),4). less(v(3),v(7),v(4)).
edge(v(4),v(5),4).
edge(v(5),v(6),4).
edge(v(1;4),v(7),4). less(v(1),v(4),v(7)).

time(0..60).
task(t(1)). task(t(1),60). tasks(t(1),t(2)).
task(t(2)). task(t(2),60). tasks(t(2),t(1)).
subtask(t(1),s(1)). subtask(t(1),s(1),v(5)).
subtask(t(1),s(2)). subtask(t(1),s(2),v(4)).
subtask(t(1),s(3)). subtask(t(1),s(3),v(2)).
subtask(t(2),s(1)). subtask(t(2),s(1),v(6)).
subtask(t(2),s(2)). subtask(t(2),s(2),v(4)).
subtask(t(2),s(3)). subtask(t(2),s(3),v(2)).

vehicle(c(1)). vehicle(c(1),v(1)).
vehicle(c(2)). vehicle(c(2),v(2)).
\end{lstlisting}

\begin{table}[t]
\centering
\begin{tabular}{|l|@{\!\!}l|}
\cline{1-2}
Atom & Meaning
\\\cline{1-2}
\mbox{\lstinline{assign(}}$c$\mbox{\lstinline{,}}$t$\mbox{\lstinline{)}}
&
$\alpha(t)=c$ for $t\in T$ and $c\in C$
\\
\mbox{\lstinline{share(}}$t$\mbox{\lstinline{,}}$t'$\mbox{\lstinline{)}}
&
$\alpha(t)=\alpha(t')$ for $\{t,t'\}\subseteq T$ such that $t<t'$ lexicographically
\\
\mbox{\lstinline{order(}}$t$\mbox{\lstinline{,}}$t'$\mbox{\lstinline{)}}
&
$t\prec t'$ for $t\in T$ and $t'\in T\setminus\{t\}$ such that $\alpha(t)=\alpha(t')$
\\
\mbox{\lstinline{check(}}$t$\mbox{\lstinline{)}}
&
$t\in T$ is topological in
$(T,\{(t',t'')\mid{}$\mbox{\lstinline{order(}}$t'$\mbox{\lstinline{,}}$t''$\mbox{\lstinline{)}} holds$\})$
\\
\mbox{\lstinline{check(}}$t$\mbox{\lstinline{,}}$t'$\mbox{\lstinline{)}}
&
$t\nprec t'$ or \mbox{\lstinline{check(}}$t$\mbox{\lstinline{)}} holds for $t\in T$ and $t'\in T\setminus\{t\}$
\\\cline{1-2}
\mbox{\lstinline{pass(}}$c$\mbox{\lstinline{,}}$t$\mbox{\lstinline{,s(}}$i$\mbox{\lstinline{),}}$v$\mbox{\lstinline{,}}$d(v)$\mbox{\lstinline{,}}$n$\mbox{\lstinline{)}}
&
$\alpha(t)=c$, $1\leq i\leq s(t)$, $t[i]=v$,
and $n\in u(c,v)\cup\{0 \mid l(c)=v\}$
\\ &
for
$t\in T$
and $c\in C$ with
$\pi(c)=\langle r_1,\dots, r_k \rangle$
such that
\\ &
$m\leq \sum_{1\leq j\leq m}d(r_j)\leq n < d(t)$,
\\ &
where
$m=\max(\{g(t[i-1])    \mid 1 < i\}\cup
        \{g(t'[s(t')]) \mid t' \prec t\})$
\\
\mbox{\lstinline{done(}}$t$\mbox{\lstinline{,s(}}$i$\mbox{\lstinline{),}}$n$\mbox{\lstinline{)}}
&
$t\in T$, $1\leq i\leq s(t)$, and $1\leq n\leq d(t)$
such that
$g(t[i])\leq {}$
\\ &
$\sum_{1\leq j\leq g(t[i])}d(r_j)\leq n$
for $\alpha(t)=c$ with
$\pi(c)=\langle r_1,\dots, r_k \rangle$
\\
\mbox{\lstinline{wait(}}$t$\mbox{\lstinline{,}}$n$\mbox{\lstinline{)}}
&
$t'\prec t$ and $0\leq n<\min\{d(t),d(t')\}$ for $t\in T$ and $t'\in T\setminus\{t\}$
\\ &
such that
$n<\max\{g(t'[s(t')]),\sum_{1\leq j\leq g(t'[s(t')])}d(r_j)\}$,
\\ &
where $\alpha(t)=\alpha(t')=c$ with
$\pi(c)=\langle r_1,\dots, r_k \rangle$
\\\cline{1-2}
\mbox{\lstinline{at(}}$c$\mbox{\lstinline{,}}$v$\mbox{\lstinline{,}}$n$\mbox{\lstinline{)}}
&
$n\in u(c,v)\cup\{0 \mid l(c)=v\}$ for $c\in C$ and $v\in V$ with
\\ &
$n\leq\max\{d(t) \mid t\in T\}$
\\
\mbox{\lstinline{move(}}$c$\mbox{\lstinline{,}}$v$\mbox{\lstinline{,}}$v'$\mbox{\lstinline{,}}$n$\mbox{\lstinline{)}}
&
$\pi(c)=\langle r_1,\dots, r_i,\dots, r_k \rangle$ for $c\in C$, $r_i\in E$, $r_i=(v,v')$,
\\ &
and $n=\sum_{1\leq j< i}d(r_j)$ with
$n+d(r_i)\leq\max\{d(t) \mid t\in T\}$
\\
\mbox{\lstinline{move(}}$c$\mbox{\lstinline{,}}$n$\mbox{\lstinline{)}}
&
\mbox{\lstinline{move(}}$c$\mbox{\lstinline{,}}$v$\mbox{\lstinline{,}}$v'$\mbox{\lstinline{,}}$n$\mbox{\lstinline{)}} holds for some connection $(v,v')\in E$
\\
\mbox{\lstinline{moving(}}$c$\mbox{\lstinline{,}}$v$\mbox{\lstinline{,}}$v'$\mbox{\lstinline{,}}$n$\mbox{\lstinline{)}}
&
$\pi(c)=\langle r_1,\dots, r_i,\dots, r_k \rangle$ for $c\in C$, $r_i\in E$, $r_i=(v,v')$,
\\ &
and $\sum_{1\leq j< i}d(r_j)<n\leq \sum_{1\leq j\leq i}d(r_j)\leq\max\{d(t) \mid t\in T\}$
\\
\mbox{\lstinline{moving(}}$v$\mbox{\lstinline{,}}$v'$\mbox{\lstinline{,}}$n$\mbox{\lstinline{)}}
&
\mbox{\lstinline{moving(}}$c$\mbox{\lstinline{,}}$v$\mbox{\lstinline{,}}$v'$\mbox{\lstinline{,}}$n$\mbox{\lstinline{)}} holds for some vehicle $c\in C$
\\
\mbox{\lstinline{free(}}$c$\mbox{\lstinline{,}}$d(v)$\mbox{\lstinline{,}}$n$\mbox{\lstinline{)}}
&
$c\in C$, $v\in h(V)\cup p(V)$, and $0\leq n\leq\max\{d(t) \mid t\in T\}$
\\ &
with $d(v)>1$ and $n'\in u(c,v)\cup\{0 \mid l(c)=v\}$ such that
\\ &
$(n'-1)\notin u(c,v)\cup\{0 \mid l(c)=v\}$, $n=n'+i*d(v)$ for some
\\ &
$i\geq 0$, and
$\{n''\in u(c,e) \mid e\in E, n'< n'' < n, {}$
\\ &
$n''+d(e)-1\leq\max\{d(t) \mid t\in T\}\} =\emptyset$
\\
\mbox{\lstinline{free(}}$c$\mbox{\lstinline{,}}$n$\mbox{\lstinline{)}}
&
$c\in C$ and $0\leq n\leq\max\{d(t) \mid t\in T\}$ such that
\\ &
$n\notin\bigcup_{v\in h(V)\cup p(V),d(v)>1}u(c,v)$
or \mbox{\lstinline{free(}}$c$\mbox{\lstinline{,}}$d$\mbox{\lstinline{,}}$n$\mbox{\lstinline{)}} holds,
\\ &
where $d=d(v)>1$ for some $v\in h(V)\cup p(V)$
\\\cline{1-2}
\mbox{\lstinline{used(}}$c$\mbox{\lstinline{,}}$n$\mbox{\lstinline{)}}
&
$c\in C$ and $0\leq n\leq\max(\{0\}\cup\bigcup_{v\in V}u(c,v))$ with
\\ &
$n\leq\max\{d(t) \mid t\in T\}$
\\
\mbox{\lstinline{move(}}$c$\mbox{\lstinline{,}}$v$\mbox{\lstinline{,}}$v'$\mbox{\lstinline{)}}
&
\mbox{\lstinline{move(}}$c$\mbox{\lstinline{,}}$v$\mbox{\lstinline{,}}$v'$\mbox{\lstinline{,}}$n$\mbox{\lstinline{)}} holds for some $0\leq n\leq\max\{d(t) \mid t\in T\}$
\\
\mbox{\lstinline{mark(}}$c$\mbox{\lstinline{,}}$v'$\mbox{\lstinline{,}}$v$\mbox{\lstinline{)}}
&
\mbox{\lstinline{move(}}$c$\mbox{\lstinline{,}}$v''$\mbox{\lstinline{,}}$v$\mbox{\lstinline{)}} holds for some connections
\\ &
$\{(v',v),(v'',v)\}\subseteq E$
such that $v'<v''$ lexicographically
\\
\mbox{\lstinline{same(}}$c$\mbox{\lstinline{,}}$c'$\mbox{\lstinline{,}}$v$\mbox{\lstinline{,}}$v'$\mbox{\lstinline{)}}
&
\mbox{\lstinline{move(}}$c$\mbox{\lstinline{,}}$v$\mbox{\lstinline{,}}$v'$\mbox{\lstinline{)}}
and
\mbox{\lstinline{move(}}$c'$\mbox{\lstinline{,}}$v$\mbox{\lstinline{,}}$v'$\mbox{\lstinline{)}}
hold for $\{c,c'\}\subseteq C$
\\ &
such that $c<c'$ lexicographically
\\\cline{1-2}
\end{tabular}
\caption{Atoms characterizing solutions to automated guided vehicle routing scenarios\label{tab:solution}}
\end{table}
Our uniform problem encoding, given in Listings~\ref{lst:encoding} and~\ref{lst:optimize},
can be understood as a merger of four logical parts,
addressing task assignment and ordering, task completion, vehicle routing, as well as
objective criteria (the latter part shown separately in Listing~\ref{lst:optimize}).
Table~\ref{tab:solution} follows this logical structure in providing the meanings
of predicates defined in each of the four parts, where a condition written in the
right column applies if and only if a corresponding atom of the form given in the left column
belongs to a stable model.

\begin{lstlisting}[label=lst:encoding,caption={Encoding of task assignment, completion, and automated guided vehicle routing},float,firstline=23,lastline=73]
% task assignment

{assign(C,T) : vehicle(C)} = 1 :- task(T).

 share(T1,T2)  :- assign(C,T1), assign(C,T2), T1 < T2.
{order(T1,T2)} :- share(T1,T2).
 order(T2,T1)  :- share(T1,T2), not order(T1,T2).

check(T1,T2) :- tasks(T1,T2), not order(T1,T2).
check(T1,T2) :- tasks(T1,T2), check(T1).
check(T2)    :- task(T2), check(T1,T2) : tasks(T1,T2).
:- task(T2), not check(T2).

% task completion

pass(C,T,s(1),V,D,N) :- assign(C,T), at(C,V,N), halt(V,D), task(T,M),
                        subtask(T,s(1),V), not wait(T,N), N < M.
pass(C,T,s(I),V,D,N) :- assign(C,T), at(C,V,N), halt(V,D), task(T,M),
                        subtask(T,s(I),V), done(T,s(I-1),N), N < M.

done(T,S,N+D) :- pass(C,T,S,V,D,N), task(T,M), not move(C,N), N+D <= M.
done(T,S,N+1) :- done(T,S,N), not task(T,N).
:- task(T,M), subtask(T,S), not done(T,S,M).

wait(T2,N) :- order(T1,T2), task(T1,M1), task(T2,M2), time(N),
              subtask(T1,s(I)), not subtask(T1,s(I+1)),
	      not done(T1,s(I),N), N < M1, N < M2.

% vehicle routing

at(C,V,0)    :- vehicle(C,V), time(0).
at(C,V2,N+D) :- move(C,V1,V2,N), edge(V1,V2,D).
at(C,V,N+1)  :- at(C,V,N), park(V,D), time(N+1), not move(C,N).
at(C,V,N+1)  :- pass(C,T,S,V,D,N), not done(T,S,N), not move(C,N).
:- node(V), time(N), #count{C : at(C,V,N)} > 1.
:- vehicle(C), time(N), #count{V : at(C,V,N)} > 1.

{move(C,V1,V2,N) : edge(V1,V2,D), time(N+D)} < 2 :- vehicle(C), time(N).
 move(C,N) :- move(C,V1,V2,N).
:- move(C,V1,V2,N), not at(C,V1,N).

moving(C,V1,V2,N+1) :- move(C,V1,V2,N).
moving(C,V1,V2,N+1) :- moving(C,V1,V2,N), time(N+1), not at(C,V2,N).
moving(V1,V2,N)     :- moving(C,V1,V2,N).
:- moving(V1,V2,N), moving(V2,V1,N).

free(C,D,N)   :- at(C,V,N), stay(V,D), not at(C,V,N-1).
free(C,D,N+D) :- free(C,D,N), time(N+D), not move(C,N).
free(C,N)     :- free(C,D,N).
free(C,N)     :- vehicle(C), time(N), not at(C,V,N) : stay(V,D).
:- move(C,N), not free(C,N).
\end{lstlisting}
In more detail, the encoding part addressing task assignment and ordering
is shown in Lines~3--12 of Listing~\ref{lst:encoding}.
The choice rule in Line~3 makes sure that exactly one atom of the form
\lstinline{assign(}$c$\lstinline{,}$t$\lstinline{)} holds per task $t\in T$,
providing the vehicle $c\in C$ such that $\alpha(t)=c$.
Further rules deal with the order~${\prec}$ among tasks
assigned to a common vehicle, where
\lstinline{share(}$t$\lstinline{,}$t'$\lstinline{)}
is derived by the rule in Line~5 if $\alpha(t)=\alpha(t')$ for tasks $t< t'$.
The choice rule in Line~6 then either picks
\lstinline{order(}$t$\lstinline{,}$t'$\lstinline{)},
expressing that $t\prec t'$,
or the rule in Line~7 yields
\lstinline{order(}$t'$\lstinline{,}$t$\lstinline{)},
standing for $t'\prec t$,
otherwise.
The remaining rules in Lines~9--12 check that the
order relation ${\prec}$ is acyclic, utilizing
modeling methods detailed in
\cite{gejari15a}.

The second encoding part in Lines~16--27 deals with the completion of transport
tasks.
To this end, the two rules in Lines~16--19 indicate time points~$n$ such that
the vehicle $c\in C$ with $\alpha(t)=c$ visits the halt node given by a subtask $t[i]$,
while any preceding subtasks according to $\prec$ as well as the sequence of subtasks
of~$t$ are already completed at time~$n$.
In case $i=1$, the latter condition is in Line~17 checked by the absence of yet
incomplete tasks $t'$ such that $t'\prec t$,
and otherwise the completion of $t[i-1]$ up to time~$n$ is required by preconditions
in Line~19.
The rule in Line~21 then derives
\lstinline{done(}$t$\lstinline{,s(}$i$\lstinline{)}\lstinline{,}$n+d(t[i])$\lstinline{)},
signaling the completion of $t[i]$,
provided that $n+d(t[i])$ does not exceed the deadline $d(t)$ for $t$ and
the vehicle~$c$ to which $t$ is assigned halts at time~$n$.
Note that the occurrence of a halt in the route of~$c$ is determined by the absence of
\lstinline{move(}$c$\lstinline{,}$n$\lstinline{)}
from a stable model,
an atom that would otherwise indicate that~$c$ takes some connection at time~$n$.
The condition that all transport tasks have to be completed within their respective deadlines
is further taken care of by the rule in Line~22, which successively propagates the
completion of a subtask $t[i]$ on to the deadline $d(t)$, along with requiring
the completion of any subtask at $d(t)$ by means of the integrity
constraint in Line~23.
Finally, the rule in Lines~25--27 derives
\lstinline{wait(}$t$\lstinline{,}$n$\lstinline{)},
an atom inspected in Line~17,
in view of a yet incomplete subtask $t'[s(t')]$ of some task $t'$ such that $t'\prec t$
at time~$n<d(t)$,
where considering the last subtask $t'[s(t')]$ of $t'$ only is sufficient because the subtasks of~$t'$
must be completed in order.

The actual routes of vehicles are tracked by the third encoding part in Lines~31--51.
To begin with, the rule in Line~31 derives
\lstinline{at(}$c$\lstinline{,}$l(c)$\lstinline{,}$0$\lstinline{)},
expressing that any vehicle $c\in C$ starts from its initial location $l(c)$ at time~$0$.
The new location $v'\in V$ resulting from a move $(v,v') \in E$ at time~$n$ is reflected by
the rule in Line~32 yielding 
\lstinline{at(}$c$\lstinline{,}$v'$\lstinline{,}$n+d((v,v'))$\lstinline{)}.
A stop at some park node $v\in p(V)$ is addressed by the rule in Line~33, deriving
\lstinline{at(}$c$\lstinline{,}$v$\lstinline{,}$n+1$\lstinline{)}
from
\lstinline{at(}$c$\lstinline{,}$v$\lstinline{,}$n$\lstinline{)}
in the absence of any move made by~$c$ at time~$n$.
Moreover, the rule in Line~34 takes care of a halt to complete some subtask $t[i]$ by
deriving
\lstinline{at(}$c$\lstinline{,}$t[i]$\lstinline{,}$n+1$\lstinline{)}
in case no subtask preceding $t[i]$ is yet incomplete and $t[i]$ itself is not
already completed at time~$n$, provided that the vehicle~$c$ with $\alpha(t)=c$
does not make any move at~$n$.
As a consequence, atoms of the form
\lstinline{at(}$c$\lstinline{,}$v$\lstinline{,}$n$\lstinline{)}
in a stable model match times $n\in u(c,v)\cup\linebreak[1]\{0 \mid l(c)=v\}$ for vehicles $c\in C$
and locations $v\in V$,
where halts can only be made for completing assigned subtasks.
Given this, the integrity constraint in Line~35 makes sure that
$u(c,v)\cap u(c',v) = \emptyset$ for any $c'\in C\setminus\{c\}$,
and the one in Line~36 asserts that $u(c,v)\cap u(c,v') = \emptyset$ for any
$v'\in V\setminus\{v\}$,
thus constituting a redundant/entailed state constraint that may still
be helpful regarding solving performance \cite{gekakasc12a}.

Unlike stops at halt or park nodes,
a move $(v,v')\in E$ by some vehicle $c\in C$ at time~$n$
such that $n+d((v,v'))$ is still of interest is signaled by an atom
\lstinline{move(}$c$\lstinline{,}$v$\lstinline{,}$v'$\lstinline{,}$n$\lstinline{)},
picked by means of the choice rule in Line~38.
The rule in Line 39 then provides the projection of such an atom to
\lstinline{move(}$c$\lstinline{,}$n$\lstinline{)},
and the integrity constraint in Line~40 requires $c$ to be
at the location~$v$ at time~$n$.
Moreover, atoms 
\lstinline{moving(}$c$\lstinline{,}$v$\lstinline{,}$v'$\lstinline{,}$n'$\lstinline{)},
derived by the rules in Lines~42--43 for $n<n'\leq n+d((v,v'))$,
yield time points $n'\in u(c,(v,v'))$.
Their projection to
\lstinline{moving(}$v$\lstinline{,}$v'$\lstinline{,}$n'$\lstinline{)}
in Line~44 is inspected by the integrity constraint in Line~45 to,
in case $(v',v)\in E$, make sure that
$u(c,(v,v'))\cap u(c',(v',v)) = \emptyset$ holds for any $c'\in C\setminus\{c\}$.
The remaining rules in Lines~47--51 check that the moves of a vehicle~$c$
respect non-atomic durations $d(v)>1$ of halt and park nodes $v\in h(V)\cup p(V)$.
To this end, an atom
\lstinline{free(}$c$\lstinline{,}$d(v)$\lstinline{,}$n$\lstinline{)}
is derived by the rule in Line~47 when $c$ is not already at~$v$ at time~$n-1$,
and then
propagated on in steps of the halt or park duration
$d(v)$ by the rule in Line~48 as long as $c$ does
not make any move.
The projection to
\lstinline{free(}$c$\lstinline{,}$n$\lstinline{)}
in Line~49 along with the rule in Line~50,
also deriving
\lstinline{free(}$c$\lstinline{,}$n$\lstinline{)}
in case $c$ is not at any
halt or park node $v\in h(V)\cup p(V)$ with $d(v)>1$ at time~$n$,
indicate all times~$n$ at which $c$ may make a move,
and the integrity constraint in Line~51 restricts moves to
such time points.

\begin{lstlisting}[label=lst:optimize,caption={Encoding part for makespan, route length, crossing, and overlap minimization},float,firstnumber=53,firstline=75,lastline=99]
% objective criteria

used(C,N)   :- at(C,V,N).
used(C,N-1) :- used(C,N), 0 < N.
:~ used(C,N). [1@4,N]
:~ used(C,N). [1@3,C,N]

move(C,V1,V2) :- move(C,V1,V2,N).

mark(C,V1,V) :- less(V1,V2,V), move(C,V2,V).
mark(C,V1,V) :- less(V1,V2,V), mark(C,V2,V).
:~ mark(C1,V1,V), move(C2,V1,V), C1 < C2. [1@2,C1,C2,V]
:~ move(C1,V1,V), mark(C2,V1,V), C1 < C2. [1@2,C1,C2,V]

same(C1,C2,V1,V2) :- move(C1,V1,V2), move(C2,V1,V2), C1 < C2.
:~ move(C1,V1,V2), move(C2,V2,V1), C1 < C2, V1 < V2. [1@1,C1,C2,V1,V2]
:~ move(C1,V2,V1), move(C2,V1,V2), C1 < C2, V1 < V2. [1@1,C1,C2,V1,V2]
:~ same(C1,C2,V1,V2), V1 < V2.                       [1@1,C1,C2,V1,V2]
:~ same(C1,C2,V2,V1), V1 < V2.                       [1@1,C1,C2,V1,V2]
:~ same(C1,C2,V1,V2), move(C1,V2,V1), V1 < V2.       [1@1,C1,C2,V2,V1]
:~ same(C1,C2,V1,V2), move(C2,V2,V1), V1 < V2.       [1@1,C1,C2,V2,V1]
:~ same(C1,C2,V2,V1), move(C1,V1,V2), V1 < V2.       [1@1,C1,C2,V2,V1]
:~ same(C1,C2,V2,V1), move(C2,V1,V2), V1 < V2.       [1@1,C1,C2,V2,V1]
:~ same(C1,C2,V1,V2), same(C1,C2,V2,V1), V1 < V2.    [2@1,C1,C2,V1,V2]
\end{lstlisting}
While the encoding parts described so far make sure that stable models match
solutions to automated guided vehicle routing scenarios,
the fourth part in Listing~\ref{lst:optimize} addresses the objective criteria
to distinguish optimal routes.
To begin with,
the rules in Lines~55--56 derive atoms
\lstinline{used(}$c$\lstinline{,}$n$\lstinline{)}
to indicate that the route of a vehicle $c\in C$ is not finished before time~$n$.
The makespan $\mathit{ms}$ is then minimized by means of the weak constraint
with the highest priority~\lstinline{4} in Line~57, associating a cost of~\lstinline{1}
with each time~$n$.
To then minimize the route length $\mathit{rl}$ with priority~\lstinline{3},
the weak constraint in Line~58 likewise penalizes
\lstinline{used(}$c$\lstinline{,}$n$\lstinline{)},
where a cost of~\lstinline{1} for each time~$n$ is charged per vehicle~$c$.
The rule in Line~60 provides the projection of moves to
\lstinline{move(}$c$\lstinline{,}$v$\lstinline{,}$v'$\lstinline{)}
in order formulate conditions regarding crossings and overlaps of routes
without referring to particular times of moves.
In fact, the rules in Lines~62--63
investigate and propagate
\lstinline{move(}$c$\lstinline{,}$v''$\lstinline{,}$v$\lstinline{)}
on to locations~$v'$ such that $(v',v)\in E$ and $v'<v''$
lexicographically,
thus signaling the inclusion of some other predecessor location of~$v$
in the route of~$c$ in terms of
\lstinline{mark(}$c$\lstinline{,}$v'$\lstinline{,}$v$\lstinline{)}.
The weak constraints in Lines~64--65 make use of this to detect a crossing
at~$v$ by the move $(v',v)$ of another vehicle $c'\in C\setminus\{c\}$,
while the lexicographical order of distinct predecessor locations of~$v$
in the routes of $c$ and $c'$ may likewise be switched.
In either case, a crossing at~$v$ accounts for a cost of~\lstinline{1}
with priority~\lstinline{2},
where the set $\{c,c'\}$ in pairs $(\{c,c'\},v)$ contributing to $\mathit{cn}$
is reflected by requiring that $c<c'$ lexicographically.
The remaining rules and weak constraints take care of overlapping connections,
where the rule in Line~67 yields
\lstinline{same(}$c$\lstinline{,}$c'$\lstinline{,}$v$\lstinline{,}$v'$\lstinline{)}
when distinct vehicles $c<c'$ both include $(v,v')$ in their routes.
This case as well as taking $(v,v')$ and $(v',v)$ 
in opposite directions
is sanctioned by a cost of~\lstinline{1} with priority~\lstinline{1}
in view of the weak constraints in Lines~68--71.
If $c$ or $c'$ includes both directions, $(v,v')$ and $(v',v)$, in its route,
an additional cost of~\lstinline{1} is charged by some of the weak
constraints in Lines~72--75,
by means of using a different term order regarding $v$ and $v'$ than before.
Finally, the weak constraint in Line~76 accounts for a cost of~\lstinline{2}
if $c$ and $c'$ both take $(v,v')$ as well as $(v',v)$,
in which case the resulting sum of four matches the number of triples 
$(\{c,c'\},e,e')$ such that $\{e,e'\}=\{(v,v'),(v',v)\}$ contributing to $\mathit{on}$.
As a consequence, we have that the sums of costs with particular priority
coincide with the objectives $\mathit{ms}$, $\mathit{rl}$,
$\mathit{cn}$, and $\mathit{on}$, where priorities follow the criteria's order of significance.

\begin{example}\label{ex:model}
The optimal solution to the automated guided vehicle routing scenario from Example~\ref{ex:instance}
indicated in Figure~\ref{fig:instance} and described further in Example~\ref{ex:solution}
is such that the vehicles $c_1$ and $c_2$ finish their last halts at $t_1[3]=t_2[3]=2$
at the times $55$ and~$49$.
Hence, a stable model representing this solution,
obtained for the facts in Listing~\ref{lst:instance} along with the encoding
in Listings~\ref{lst:encoding} and~\ref{lst:optimize},
contains the atoms
\lstinline{at(c(1),v(2),55)} and \lstinline{at(c(2),v(2),49)}.
These in turn imply atoms
\lstinline{used(c(}$i$\lstinline{),}$n_i$\lstinline{)} 
for $i\in\{1,2\}$, $0\leq n_1\leq 55$, and $0\leq n_2\leq 49$.
The weak constraints with priorities~\lstinline{4} and~\lstinline{3} thus yield
$56$ or $106$ distinct terms, respectively, of the form
\lstinline{[1@4,}$n_i$\lstinline{]} or
\lstinline{[1@3,c(}$i$\lstinline{),}$n_i$\lstinline{]},
where the numbers of terms match $\mathit{ms}+1=55+1=56$ and $\mathit{rl}+i=104+2=106$.
Note that the addends~$1$ or~$i$ to $\mathit{ms}$ and $\mathit{rl}$ are due to atoms
of the form \lstinline{used(c(}$i$\lstinline{),0)},
which can be deterministically derived in view of the initial locations of vehicles
and do not affect the preference order of solutions.
Regarding crossings, as detailed in Example~\ref{ex:optimal},
the connections taken by $c_1$ and $c_2$ trigger weak constraints
associated with the terms
\lstinline{[1@2,}\linebreak[1]\lstinline{c(1),}\linebreak[1]\lstinline{c(2),}\linebreak[1]\lstinline{v(1)]},
\lstinline{[1@2,}\linebreak[1]\lstinline{c(1),}\linebreak[1]\lstinline{c(2),}\linebreak[1]\lstinline{v(4)]}, and
\lstinline{[1@2,}\linebreak[1]\lstinline{c(1),}\linebreak[1]\lstinline{c(2),}\linebreak[1]\lstinline{v(7)]},
leading to the sum $\mathit{cn}=3$ of costs.
Moreover, six unidirectional overlaps of the routes of $c_1$ and $c_2$
are indicated by terms of the form
\lstinline{[1@1,c(1),c(2),v(}$v$\lstinline{),v(}$v'$\lstinline{)]} for
$(v,v')\in \{($%
\lstinline{1}$,$\lstinline{2}$),\linebreak[1]($%
\lstinline{2}$,$\lstinline{3}$),\linebreak[1]($%
\lstinline{3}$,$\lstinline{4}$),\linebreak[1]($%
\lstinline{4}$,$\lstinline{5}$),\linebreak[1]($%
\lstinline{5}$,$\lstinline{6}$),\linebreak[1]($%
\lstinline{1}$,$\lstinline{6}$)\}$,
among which the lexicographically ordered pair $($\lstinline{1}$,$\lstinline{6}$)$
stands for the connection $(6,1)$ taken in the opposite direction.
However, such term ordering matters for bidirectional connections only,
where $(v,v')\in \{($%
\lstinline{1}$,$\lstinline{4}$),\linebreak[1]($%
\lstinline{4}$,$\lstinline{7}$)\}$
yield three distinct terms each:
\lstinline{[1@1,}\linebreak[1]\lstinline{c(1),}\linebreak[1]\lstinline{c(2),}\linebreak[1]\lstinline{v(}$v$\lstinline{),}\linebreak[1]\lstinline{v(}$v'$\lstinline{)]},
\lstinline{[1@1,}\linebreak[1]\lstinline{c(1),}\linebreak[1]\lstinline{c(2),}\linebreak[1]\lstinline{v(}$v'$\lstinline{),}\linebreak[1]\lstinline{v(}$v$\lstinline{)]}, and
\lstinline{[2@1,}\linebreak[1]\lstinline{c(1),}\linebreak[1]\lstinline{c(2),}\linebreak[1]\lstinline{v(}$v$\lstinline{),}\linebreak[1]\lstinline{v(}$v'$\lstinline{)]}.
That is, the weak constraints with priority~\lstinline{1} lead to twelve terms in total,
ten of them accounting for a cost of~\lstinline{1} and two having a cost of~\lstinline{2} (written before `\lstinline{@}'),
summing up to $\mathit{on}=14$.
\eofex
\end{example}


\section{Empirical Results}\label{sec:experiments}

A preliminary field study by production engineers at Mercedes-Benz Ludwigsfelde GmbH
investigated viable options for implementing a new control system for
automated guided vehicles used in car assembly,
where one goal consists of setting the routes for such vehicles up automatically rather than by hand.
This study raised interest in the
open-source control system software \openTCS\ (see \url{http://www.opentcs.org}), 
developed by Fraunhofer IML. 
While \openTCS\ offers a generic platform for automated guided vehicle control,
it is not geared to ship a solver for hard combinatorial tasks such as
the multi-objective optimization of routes off the shelf.
Instead, the system is extensible and allows for integrating customized components,
which is where our ASP approach to vehicle routing is envisaged to come into play in the future.
For reference, however, we contrast our results to those obtained with the
default scheduler of \openTCS, which implements a simple round-robin procedure
to assign tasks to and pick shortest routes for vehicles.

In order to evaluate the practical applicability of control systems,
the engineers at Mercedes-Benz Ludwigsfelde GmbH defined 18 use cases
based on a factory layout with 25 locations and 35 connections, 
resembling the storage areas, assembly lines, and transport routes at the
physical car factory.
Such use cases are run in simulation to test the entire functionality
of a control system, where task assignment and vehicle routing are incorporated
at the high level.
According to the main test targets,
the use cases %
(available at \url{http://www.cs.uni-potsdam.de/wv/projects/daimler/resources-iclp18.tar.xz})
are grouped into five categories:
A) communication and feedback,
B) task assignment and execution,
C) routing and traffic management,
D) special conditions and breakdowns, and
E) full production cycle.
While test cases in the first four groups focus on particular scenarios of 
small size, i.e., up to three tasks and vehicles, the use case in category~E emulates 
a full production cycle of roughly 20 minutes in real time,
involving ten transport tasks with 39 subtasks in total to be accomplished 
by four automated guided vehicles.

%
\begin{figure}[t]
\centering
\includegraphics[scale=0.35]%
{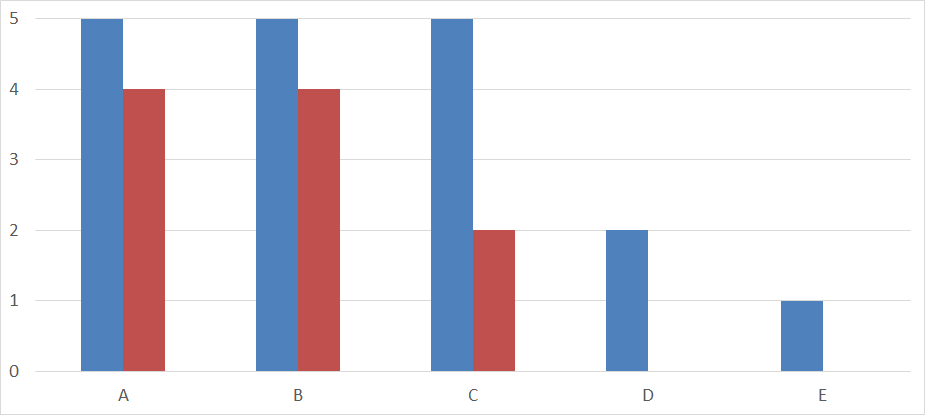}
\caption{Use cases solved by means of ASP (in blue) and the
         default scheduler of \openTCS\ (in red)\label{fig:chart}}
\end{figure}
For computing optimal routes with \clingo\ (version 5.2.2), we represented the scenarios
of the use cases in terms of facts as described in the previous section,
thus focusing on task assignment and vehicle routing rather than system control.
As indicated by the blue bars in Figure~\ref{fig:chart},
we succeeded in obtaining provably optimal routes in all of the 18 use cases,
where the runtimes of \clingo\ on a desktop machine equipped with Intel i7-6700 3.40GHz CPU
ranged from split seconds to ten seconds at most for each of the 17 small scenarios
in the first four groups.
While these small instances stem from use cases primarily created to
test the reaction of a control system to some isolated situation,
the scenario in~E aims at managing all of the transport tasks
recurring within each full production cycle, a complex operation
that has so far been carried out by human engineers before a production
process starts or is resumed with an updated configuration, respectively.
In fact, optimizing the routes for automated guided vehicles over the full
production cycle is computationally challenging, even for a system like
\clingo\ that is geared to hard combinatorial problem solving,
and it took about 5 hours to compute a provably optimal solution.
Its makespan of $225$ clock cycles amounts to five times as many seconds
in real time, roughly 20 minutes,
the route length $891$ corresponds to about 75 minutes of operation,
shared between four vehicles whose computed routes involve $39$ crossings
and $141$ overlaps in total.
While 5 hours of computation time can certainly not be afforded by a
control system that has to react to incidents in real time,
the dimension of what has to be anticipated in comparable scenarios
is nevertheless encouraging, as it tells us that routes for accomplishing 
recurrent transport tasks in realistic production processes can be programmed
in an automated fashion, going along with optimality that cannot be
established by human engineers alone.

The red bars shown in Figure~\ref{fig:chart} exhibit that the default scheduler
shipped with \openTCS\ fails to come up with a feasible solution in 8 of the 18
use cases, which is due to its greedy approach to assign a task and lock locations along the
shortest route of a vehicle one by one.
E.g., such an approach is bound to fail on the example scenario in Figure~\ref{fig:instance},
where each of the vehicles $c_1$ and $c_2$ must necessarily visit node~$4$ in order
to complete any subtask, which locks the respective other vehicle out from visiting node~$4$,
while a single vehicle cannot alone complete both transport tasks, $t_1$ and $t_2$, within their deadlines.
This observation along with the fact that \clingo\ can timely handle the small scenarios
of use cases in the first four groups clearly motivate its integration into \openTCS\
as a component in charge of task assignment and vehicle routing,
which we are working on.


\section{Discussion}\label{sec:discussion}

The problem of assigning
transport tasks to and routing automated guided vehicles in car assembly
reveals parallels to Generalized Target Assignment and Path Finding (GTAPF;~\citeNP{ngobsoscye17a})
as well as Temporal Planning (cf.\ \citeNP{fisher08a}).
Similar to automated guided vehicle routing scenarios, GTAPF allows for more tasks than there are agents,
sequences of subtasks, and deadlines for tasks,
while its framework assumes that tasks are completed according to some total order and it does
not feature (non-atomic) durations.
Unlike that, durations play a fundamental role in Temporal Planning,
but standard representations such as the Planning Domain Definition Language (PDDL;~\citeNP{foxlon03a})
lack constructs for conveniently modeling task assignment and ordering.
We thus stick to a native ASP encoding, and a corresponding approach likewise turned
as advantageous for planning by means of tabled logic programming
\cite{zhbado15a}.

Early work on using ASP for solving basic multi-agent path finding problems was done by \citeN{erkiozsc13a};
also, \citeN{ngobsoscye17a} used ASP for solving GTAPF problems.
Action durations and intervals were previously considered in ASP by \citeN{sobatu04a}.
Notably, we checked that \clingo\ requires less than 1~GB of RAM
to represent durations and makespan of a full production cycle
at the car factory of Mercedes-Benz Ludwigsfelde GmbH,
while extensions by difference logic
\cite{newascha17a}
or integer variables
\cite{bakaossc16a}
may possibly handle even greater time intervals in a compact way.

The contributions of our work include a transparent and elaboration tolerant
formalization of transport tasks evolving in realistic production processes
by specifying them in ASP,
thus also making off-the-shelf solving systems accessible for computing optimal vehicle routes.
As it turns out, our declarative approach allows for handling a
scenario covering the full production cycle at the car factory of
Mercedes-Benz Ludwigsfelde GmbH,
so that it can assist human engineers in setting up routes to
be periodically taken by automated guided vehicles.
Moreover, scenarios of small size, resembling task reassignment or vehicle rerouting in case of unforeseen
circumstances, can virtually be handled in real time,
which also makes the future integration of \clingo\
as a component of the control system software \openTCS\
for task assignment and vehicle routing
attractive.

%
%


\vspace{-7.25pt}
\paragraph{Acknowledgments}

This work was partially funded by DFG grant SCHA 550/9.
We are grateful to the anonymous reviewers for their helpful comments.






\begin{thebibliography}{}\preduce

\bibitem[\protect\citeauthoryear{Banbara, Kaufmann, Ostrowski, and
  Schaub}{Banbara et~al\mbox{.}}{2017}]{bakaossc16a}
{\sc Banbara, M.}, {\sc Kaufmann, B.}, {\sc Ostrowski, M.}, {\sc and} {\sc
  Schaub, T.} 2017.
\newblock Clingcon: The next generation.
\newblock {\em Theory and Practice of Logic Programming\/}~{\em 17,\/}~4,
  408--461.

\bibitem[\protect\citeauthoryear{Calimeri, Faber, Gebser, Ianni, Kaminski, Krennwallner, Leone, Ricca, and Schaub}{Calimeri et~al\mbox{.}}{2012}]{aspcore2}
{\sc Calimeri, F.}, {\sc Faber, W.}, {\sc Gebser, M.}, {\sc Ianni, G.}, {\sc Kaminski, R.}, {\sc Krennwallner, T.}, {\sc Leone, N.}, {\sc Ricca, F.}, {\sc and} {\sc Schaub, T.} 2012.
\newblock {ASP-Core-2}: Input language format.

\bibitem[\protect\citeauthoryear{Erdem, Kisa, {\"{O}}ztok, and Sch{\"{u}}ller}{Erdem et~al\mbox{.}}{2013}]{erkiozsc13a}
{\sc Erdem, E.}, {\sc Kisa, D.}, {\sc {\"{O}}ztok, U.}, {\sc and} {\sc Sch{\"{u}}ller, P.} 2013.
\newblock A general formal framework for pathfinding problems with multiple agents.
\newblock In {\em Proceedings of AAAI'13}. 
AAAI Press, 290--296.

\bibitem[\protect\citeauthoryear{Fisher}{Fisher}{2008}]{fisher08a}
{\sc Fisher, M.} 2008.
\newblock Temporal representation and reasoning.
\newblock In {\em Handbook of Knowledge Representation}. Elsevier Science, 513--550.

\bibitem[\protect\citeauthoryear{Fox and Long}{Fox and Long}{2003}]{foxlon03a}
{\sc Fox, M.} {\sc and} {\sc Long, D.} 2003.
\newblock {PDDL}2.1: An extension to {PDDL} for expressing temporal planning domains.
\newblock {\em Journal of Artificial Intelligence Research\/}~{\em 20}, 61--124.

\bibitem[\protect\citeauthoryear{Gebser, Janhunen, and Rintanen}{Gebser et~al\mbox{.}\ignorespaces}{}
]{gejari15a}
{\sc Gebser, M.}, {\sc Janhunen, T.}, {\sc and} {\sc Rintanen, J.} 
\newblock Declarative encodings of acyclicity properties.
\newblock {\em Journal of Logic and Computation,\/}
in press.

\bibitem[\protect\citeauthoryear{Gebser, Kaminski, Kaufmann, Lindauer, Ostrowski, Romero, Schaub, and Thiele}{Gebser et~al\mbox{.}}{2015}]{PotasscoUserGuide}
{\sc Gebser, M.}, {\sc Kaminski, R.}, {\sc Kaufmann, B.}, {\sc Lindauer, M.}, {\sc Ostrowski, M.}, {\sc Romero, J.}, {\sc Schaub, T.}, {\sc and} {\sc Thiele, S.} 2015.
\newblock {\em Potassco User Guide}.
\newblock University of Potsdam.

\bibitem[\protect\citeauthoryear{Gebser, Kaminski, Kaufmann, and Schaub}{Gebser et~al\mbox{.}}{2012}]{gekakasc12a}
{\sc Gebser, M.}, {\sc Kaminski, R.}, {\sc Kaufmann, B.}, {\sc and} {\sc Schaub, T.} 2012.
\newblock {\em Answer Set Solving in Practice}.
\newblock Morgan and Claypool Publishers.

\bibitem[\protect\citeauthoryear{Lifschitz}{Lifschitz}{1999}]{lifschitz99b}
{\sc Lifschitz, V.} 1999.
\newblock Answer set planning.
\newblock In {\em Proceedings of ICLP'99}. 
MIT Press, 23--37.

\bibitem[\protect\citeauthoryear{Neubauer, Wanko, Schaub, and Haubelt}{Neubauer
  et~al\mbox{.}}{2017}]{newascha17a}
{\sc Neubauer, K.}, {\sc Wanko, P.}, {\sc Schaub, T.}, {\sc and} {\sc Haubelt,
  C.} 2017.
\newblock Enhancing symbolic system synthesis through {ASPmT} with partial
  assignment evaluation.
\newblock In {\em Proceedings of DATE'17}. 
  IEEE Press, 306--309.

\bibitem[\protect\citeauthoryear{Nguyen, Obermeier, Son, Schaub, and Yeoh}{Nguyen et~al\mbox{.}}{2017}]{ngobsoscye17a}
{\sc Nguyen, V.}, {\sc Obermeier, P.}, {\sc Son, T.}, {\sc Schaub, T.}, {\sc and} {\sc Yeoh, W.} 2017.
\newblock Generalized target assignment and path finding using answer set programming.
\newblock In {\em Proceedings of IJCAI'17}. 
  IJCAI/AAAI Press, 1216--1223.

\bibitem[\protect\citeauthoryear{Son, Baral, and Tuan}{Son et~al\mbox{.}}{2004}]{sobatu04a}
{\sc Son, T.}, {\sc Baral, C.}, {\sc and} {\sc Tuan, L.} 2004.
\newblock Adding time and intervals to procedural and hierarchical control specifications.
\newblock In {\em Proceedings of AAAI'04}. 
  AAAI Press, 92--97.

\bibitem[\protect\citeauthoryear{Zhou, Bart{\'a}k, and Dovier}{Zhou
  et~al\mbox{.}}{2015}]{zhbado15a}
{\sc Zhou, N.}, {\sc Bart{\'a}k, R.}, {\sc and} {\sc Dovier, A.} 2015.
\newblock Planning as tabled logic programming.
\newblock {\em Theory and Practice of Logic Programming\/}~{\em 15,\/}~4-5,
  543--558.

\end{thebibliography}



\end{document}